\documentclass[review]{elsarticle}

\usepackage{hyperref}
\usepackage{pifont}
\usepackage{booktabs}
\newcommand{\cmark}{\ding{51}}
\newcommand{\xmark}{\ding{55}}
\usepackage{makecell}
\usepackage{float}

\usepackage{amsmath}
\usepackage{soul} 
\usepackage{xcolor} 

\journal{Journal of \LaTeX\ Templates}

\begin{document}

\begin{frontmatter}

\title{RGH-GNN : Risk-Gated Heterogeneous Graph Neural Network for Financial Fraud Detection}

\author[1]{Mythreyee Hari}
\corref{cor1}
\ead{mythreyeeh2023@gmail.com}
\author[1]{Subin Sahayam}
\author[1]{Vegesna S M Srinivasavarma}

\address[1]{Department of Computer Science and Engineering, Shiv Nadar University, Chennai - 603110, Tamil Nadu, India}

\cortext[cor1]{Corresponding author}
\begin{abstract}
Credit card fraud is a significant problem for financial institutions, resulting in substantial monetary loss each year. Fraud patterns and methods are continuously evolving, with fraudulent transactions often disguised to resemble legitimate activity. Real world transaction data for fraud detection is severely affected by class imbalance making it harder to detect these transactions. Fraudulent transactions are rarely isolated. Accounts, devices, and other common attributes are often shared across fraudulent activity, revealing relational patterns that individual transaction features alone cannot capture. Graph-based approaches model these relationships explicitly. However, standard graph neural networks aggregate neighbourhood information uniformly, which can dilute the already scarce signal carried by fraudulent transactions
and limit detection of camouflaged fraud. This work presents RGH-GNN, a heterogeneous graph neural network that addresses this limitation by incorporating transaction-level risk scores, generated by a supervised LightGBM classifier as edge weights. It is used to guide message passing through a learnable risk-gated aggregation mechanism. It ensures that high-risk transactions contribute more strongly to learned representations rather than being diluted by uniform neighbourhood aggregation. RGH-GNN is evaluated on the IEEE-CIS fraud detection dataset against standard graph neural network models that are trained on the same heterogeneous graph. RGH-GNN outperforms all baselines across every evaluation metric, with an AUC-ROC score of 0.914. This demonstrates that incorporating engineered risk signals into graph-based message passing substantially improves fraud detection performance. 
\end{abstract}


\end{frontmatter}


\section{Introduction}
Financial transactions form the backbone of modern digital economies, enabling the legal exchange of assets between individuals and enterprises. It has led to a rapid growth in online financial platforms which has been accompanied by an increase in financial fraud. It can be defined as the unlawful acquisition or manipulation of assets. Among various forms of fraud, credit card fraud constitutes a significant portion, with millions of cards compromised annually and substantial financial losses incurred worldwide \cite{global_credit_card_statistics}.

A key characteristic of financial fraud is that it rarely occurs as an isolated event. Fraudulent activities are often carried out in coordinated patterns across multiple transactions, involving shared accounts, devices, or identities, and often occur during unusual temporal windows \cite{relational_fraud}. This interconnected nature makes fraud inherently relational. Suspicious behaviour is revealed as a collection of transactions and their interactions rather than just individual transactional features.

Traditional fraud detection systems rely on manual audits, rule-based mechanisms, and statistical techniques \cite{traditional_fraud_detection}. These are effective against specific static fraud patterns but limited due to their inability to adapt to other evolving fraud strategies. More recently, machine learning models that are trained on tabular transaction data have improved detection performance by learning patterns from historical data \cite{ml_finanial_fraud}. These models treat each transaction independently, relying solely on the features in a single row. However, they fail to capture the underlying relationships between different transactions and entities within the dataset.

These models also suffer from other key challenges including class imbalance, camouflage, and the heterogeneous nature of financial systems. The occurrence of fraud in financial datasets is scarce, which results in severe class imbalance. This makes it difficult for machine learning models to learn accurate representations of fraud and flag it reliably. Techniques to address this imbalance must be incorporated during preprocessing and training. The camouflage problem is a challenge where fraudsters deliberately share attributes, such as IP addresses, devices, or merchants, with legitimate transactions in order to blend their behavior with normal ones \cite{care_gnn}. Such transactions are difficult to flag when inspected independently, since they appear normal in isolation. Financial systems also contain multiple entities like accounts, transactions, cards, IP addresses, and merchants etc. Each of these entities contribute distinct information relevant to fraud detection. The attributes shared within entity types, and the connections formed across different entity types, need to be accurately represented with their independent feature sets. Capturing this structure requires modeling financial systems as a heterogeneous network rather than a tabular structure in order to effectively detect fraud.

To address these limitations, graph-based solutions model financial systems as networks. Transactions, accounts, and other financial entities are represented as nodes connected through meaningful relationships \cite{care_gnn, gnn_fraud_detection_other}. It allows the incorporation of relational context, letting the model detect patterns such as shared credentials or coordinated activity. 

Existing graph neural networks that rely on uniform aggregation of neighbouring nodes face a critical challenge. Aggregation strategies that provide equal weightage to both normal and fraudulant nodes can dilute important fraud signals. This problem is particularly severe in highly imbalanced datasets where fraudulent transactions are rare \cite{care_gnn}. More recently, ensemble methods have demonstrated better overall performance by combining multiple perspectives such as tabular and relational views of data \cite{ensemble}. Nevertheless, effectively integrating these views while preserving high-risk signals remains an open challenge.

Motivated by these limitations, this work proposes Risk-Gated Heterogeneous Graph Neural Network (RGH-GNN), a risk-aware graph-based framework for fraud detection that integrates domain-driven feature engineering, tabular risk modeling, and adaptive graph learning. We use a gradient boosting model, which excels on tabular transaction features, to generate transaction-level risk scores. These scores are used during graph message passing to guide the aggregation of neighboring nodes based on their risk to flag the most suspicious transactions. The proposed framework is evaluated on the IEEE-CIS fraud detection dataset \cite{ieee_cis}, a widely used benchmark for credit card fraud detection. It contains real-world transaction records with associated identity and card information. The experiments evaluated RGH-GNN's performence across multiple metrics and revealed that it provides a reliable framework for generating and utilizing risk scores in such a way that they could provide meaningful signals for fraud, which most baselines struggled with.

The main contributions of the work are as follows:

\begin{itemize}
\item \textbf{Domain-driven feature engineering:}  Feature engineering strategies that have been identified in literature are engineered on the dataset. Temporal patterns, transaction dynamics, and behavioral signals that are relevant for fraud detection are introduced for both transaction and account node features.
\item \textbf{Heterogeneous graph construction:} The engineered features are used to construct nodes of the graph that represent different entities and their relevant attributes. Relationships between transactions and accounts, as well as other entities of the graph are represented by edges.
\item \textbf{Integration of tabular risk modeling:} A gradient boosting model (LightGBM) is used to generate informative risk scores to guide graph learning. This risk score is added to the account-transaction edges constructed in the graph.
\item \textbf{Risk-aware graph neural network (RGH-GNN):} A novel learnable gating strategy uses the risk scores engineered on the graph to decide the confidence with which neighbours of the target node are selected during message passing. It is used to handle class imbalance and reduces signal dilution.
\item \textbf{Evaluation :} RGH-GNN, which utilizes the novel gating strategy for a risk-aware perspective of the graph, is tested across multiple metrics. It outperforms all the baselines with an AUC-ROC score of 0.914, demonstrating the model's ability to flag fraud reliably across multiple thresholds. 
\end{itemize}

The rest of the paper is structured as follows : Section~\ref{sec:related_work} discusses related works and their limitations that serve as the motivation for this research. Section~\ref{sec:methodology} describes the methodology and components of the RGH-GNN architecture. Section~\ref{sec:experiments} describes the training process, metrics, ablation, baseline comparison and performance of RGH-GNN against other algorithms. Section~\ref{sec:discussion} discusses the strengths and limitations of the proposed architecture. Section~\ref{sec:conclusion} concludes the work and suggests future directions to extend this research.

\section{Related Work}
\label{sec:related_work}
Following the limitations of static rule-based systems, machine learning approaches were introduced for the classification of fraudulent transactions. Boosting algorithms like XGBoost have demonstrated strong performance on structured transaction data by iteratively correcting the errors of weak learners \cite{xgboost}. Isolation Forest has been applied as an unsupervised classifier for learning patterns in cases of scarcity of labeled fraud \cite{isolation_forest, isolation_forest_2}. Clustering-based models such as K-means have been used to group transactions by behavioural similarity, and flag outliers as fraud \cite{kmeans_clustering}. SMOTE-based oversampling has been combined with classifiers to address the severe class imbalance in fraud datasets \cite{smote_comparative}. Machine learning approaches rely on extensive manual feature engineering which is difficult to perform on high-dimensional transaction data. They also treat transactions as independent samples, which makes it difficult to capture non-linear and relational dependencies in a financial system.

Deep learning approaches were subsequently introduced to overcome the limitations of machine learning models. Convolutional neural networks have been used to extract local patterns from transaction feature sequences to classify fraud \cite{cnn}. Recurrent neural network architectures, particularly LSTMs, have allowed temporal features in transactions to be captured for sequential fraud detection \cite{lstm, autoencoder_lstm}. Autoencoder-based approaches work on the principle of reconstructing transaction patterns from the dataset and flagging high reconstruction error as fraudulent \cite{autoencoder_lstm, autoencoder}. GANs have been used to generate synthetic fraudulent samples to solve the data sparsity and class imbalance problem in fraud datasets \cite{gan}. More recently, transformer-based architectures which leverage self-attention have been used to capture long-range dependencies across transactions without the sequential bottleneck of recurrent models \cite{transformer_fintech}. Deep learning architectures are capable of automatically learning feature representations without manual engineering. However, they face the same limitation of treating transactions as independent samples. They cannot capture relationships between transactions and miss out on the structural patterns of coordinated fraud such as fraud rings and chain transactions.

Graph-based approaches overcome the limitation of non-relational representation by modeling transaction systems as networks \cite{gnn_review}. Foundational GNN architectures include GraphSAGE \cite{graphsage}, Graph Convolutional Networks (GCN) \cite{gcn} and Graph Attention Networks (GAT) \cite{gat}. These architectures are designed for homogeneous graphs, where all nodes and edges are treated as a single type. It limits their ability to represent different features unique to different kinds of entities such as accounts, transactions, and devices.
Heterogeneous graph representations address this limitation by modeling graphs with multiple distinct entity types as nodes. These nodes are connected by edges that describe different relationships between them. Common heterogeneous architectures include Heterogeneous Graph Attention Network (HAN) \cite{han} and Heterogeneous Graph Transformer (HGT) \cite{hgt}. Heterogeneous networks are suitable for fraud detection because coordinated fraud can be found across multiple entities including transactions and accounts.

In supply chain finance, heterogeneous graph neural networks have been used to capture features specific to different entity types, such as enterprises and transactions \cite{heterogeneous_supply_chain}. GNNs have also utilized local augmentation and adaptive relation aggregation methods on multi-relational graphs to uncover feature and relation camouflage.\cite{gnn-laara}. While heterogeneous graph representations preserve entity-specific semantics, they still face some crucial challenges. First, financial transaction graphs are sparse and contain noisy edges due to the quality and construction of the graph \cite{calibrated_risk_structural}. Second, fraudsters actively employ evolving camouflage strategies to disguise suspicious transaction patterns \cite{global_confidence_gnn}. Finally, uniform aggregation strategies can dilute informative risk signals and increase noise from irrelevant nodes. CARE-GNN extends fraud detection methodologies using label-aware similarity measures and reinforcement learning to select informative neighbours which mitigates the camouflage behaviour of fraudsters \cite{care_gnn}. A temporal-semantic dynamic quantifier (TSDQ) \cite{temporal_semantic} is used in one proposed architecture, to capture temporal and semantic information from transaction nodes. This can help with understanding transaction behaviour and reveal potential camouflaged fraudsters.

Recent work has also explored incorporating risk signals into the graph learning process to guide message passing. One recent methodology suggests a transformer and encoder architecture to represent relations between transactions and capture their semantics \cite{hif-fd}. It further utilizes an attention mechanism and proportionally models relational risk propagation with edge-aware attention. Supply-chain systems utilize risk aware frameworks by merging heterogeneous data streams and incorporating multi-head attention mechanisms \cite{supply_chain_risk}. These are used to assign dynamic coefficients to risk-sensitive features for more robust fraud detection. Similarly, blockchain systems utilize SMOTE and noise augmentation on data, followed by a novel risk-aware scoring mechanism to prioritize high-risk fraud detection \cite{blockchain_risk}. Risk diffusion mechanisms have modeled how risk propagates across financial networks \cite{finguard_gnn}. It uses multiple methods to capture varying influence patterns between suspicious and normal nodes. Risk-aware orthogonal disentanglement is a semi-supervised approach that uses label-derived risk encoding to project aggregated node representations into separate subspaces for camouflage relevant fraud classification \cite{risk_aware_disentanglement}.

These works establish that incorporating risk-aware signals into graph-based fraud detection improves robustness to camouflage, noise, and class imbalance. 
Risk-based frameworks must not only derive meaningful risk scores but also translate them into effective signals for aggregation strategies. While some existing architectures may assign risk scores to transactions, their mechanisms for incorporating these scores into message passing to provide meaningful fraud signals can be limited. RGH-GNN addresses both these aspects. It uses an independently trained gradient boosting model to produce calibrated, transaction-level risk scores. During graph construction, these scores are used directly as edge weights to guide message passing in a heterogeneous account-transaction graph. A learnable risk-gated aggregation mechanism is used during message passing to leverage these risk scores and flag fraud.

Some of the most popular datasets found in credit card fraud detection literature include IEEE-CIS Fraud Detection \cite{ieee_cis}, Credit Card Fraud Detection \cite{ccfd}, and Fraud ecommerce \cite{fraudecom}. All of these datasets are used to identify transaction fraud among a set of credit card transactions. PaySim \cite{paysim} is a financial simulator that is used to generate synthetic mobile transaction datasets that models are trained on to identify malicious behavior. D-graph \cite{dgraph} is a large real-world dynamic graph dataset built for financial anomaly detection. It contains about 3M nodes, 4M dynamic edges, and 1M ground-truth nodes. The Yelp-Chi and Amazon datasets are also benchmark anomaly detection datasets that have been used for evaluating GNN performance \cite{care_gnn}. Yelp-Chi comprises hotel reviews from the Yelp platform, labeled as legitimate or spam, and Amazon contains e-commerce product reviews for a range of orders placed on the Amazon platform. Multiple other fraud detection datasets are used in literature to measure the performance of anomaly detection models based on the domain and classification type for fraud \cite{dataset_benchmarks}. The IEEE-CIS dataset was chosen for the experiments of RGH-GNN since it is widely utilized for the credit card fraud detection problem. It contains extensive features for both transactions and identities which can be used to engineer a highly informative graph to test the performance of this algorithm. 

\section{Methodology}
\label{sec:methodology}
\begin{figure}[ht]
    \centering
    \includegraphics[width=1\textwidth]{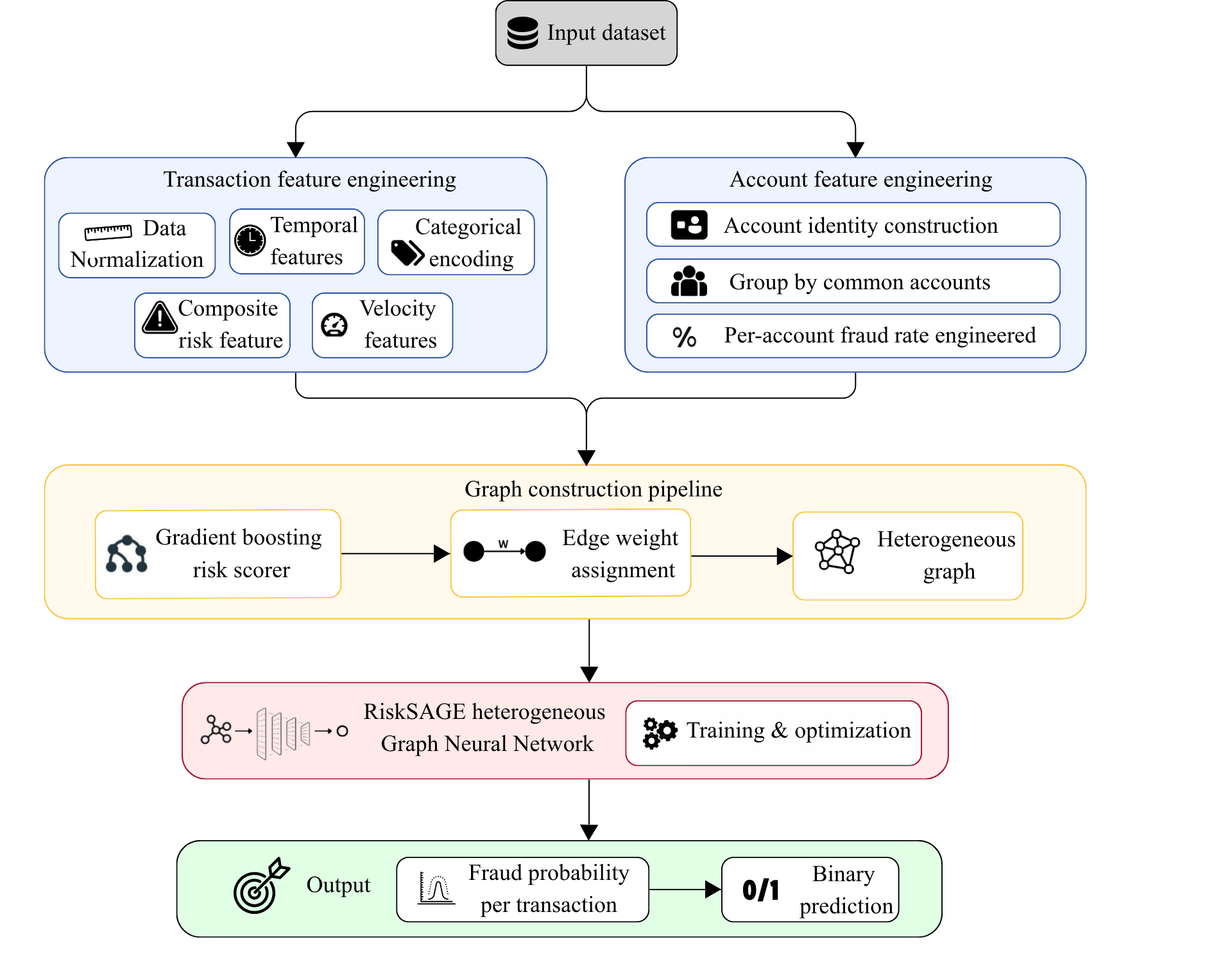}
    \caption{Methodology Diagram RGH-GNN Pipeline}
    \label{fig:methodologyDiagram}
\end{figure}

Fraudulent transactions involve a network of shared accounts, devices, and behavioral patterns that collectively signal risk. RGH-GNN exploits this structure by representing the dataset as a heterogeneous bipartite graph, with two distinct node types, namely accounts and transactions, which are connected by different types of edges based on shared identity. Figure~\ref{fig:methodologyDiagram} presents the overall architecture of the proposed RGH-GNN framework, which includes feature engineering, heterogeneous graph construction, graph neural network training, and transaction-level fraud prediction.
\subsection{Constructing the Heterogeneous Transaction Graph}
\label{sec:graph_construction}
In practice, preprocessing and feature engineering strategies vary across datasets depending on the entities, features, and relational structure present. For every dataset, the feature selection strategy can be defined as 
\[
D = \{f_1, f_2, \ldots, f_n\}
\]

where $D$ is the set of all $n$ features in the dataset. Feature selection
involves selecting a subset $D' \subset D$ containing $m$ features, such that

\[
D' = \{f_{i_1}, f_{i_2}, \ldots, f_{i_m}\},
\qquad m < n.
\]
RGH-GNN requires a heterogeneous graph as input. It should contain distinct entity types that are represented as separate nodes connected by edges, where at least one edge type is designated to carry risk scores as weights during message passing. The graph construction strategy can therefore vary across datasets, provided the resulting graph captures the relational structure of the transaction network in this form. 
The preprocessing and feature engineering strategies described in section ~\ref{sec:graph_construction} are for features commonly present in financial transaction datasets and serve as a reference implementation of this stage within the RGH-GNN pipeline.
\subsubsection{Feature Engineering}
This section describes feature engineering strategies for common transaction and account features separately:
\begin{enumerate}

\item\textbf{Transaction-Level Features:}
\begin{enumerate}

\item{\textbf{Transaction Amount: }}
Transaction amount describes how much money is transferred during a transaction. It is log-transformed to compress the heavy-tailed distribution of amount values, making unusually high transaction amounts easier to detect as important signals for fraud.
\begin{equation}
f_{\text{amt}} = \log(1 + T)
\end{equation}
where $T$ is the transaction amount.

\item{\textbf{Temporal Features: }}
The transaction timestamp, which is often measured in seconds, is converted in terms of hour of the day. It captures temporal windows when transactions occur, since suspicious transactions in low-activity windows can serve as an important signal for fraud.
\begin{equation}
f_{\text{hour}} = \frac{D \bmod (n \cdot i)}{i}
\end{equation}
where $D$ is the transaction timestamp in seconds, $i$ is the number of seconds in a time interval, and $n$ is the number of time intervals in one day (e.g., we assume $i = 3600$, since 3600 seconds make up one hour. We assume $n = 24$, since 24 hours make up a day).

\item{\textbf{Behavioral Gap Features: }}
Features that represent the time elapsed since prior account activity are clipped to remove negative values and log-transformed to compress their heavy-tailed distribution. These help reveal abnormally rapid or repeated activity that is often indicative of fraud.
\begin{equation}
f_{D_k} = \log\bigl(1 + \max(D_k, 0)\bigr), \quad k = 1, \ldots, K
\end{equation}
where $D_k$ is the $k$-th delta-time feature and $K$ is the total number of delta-time features available in the dataset.

\item{\textbf{Velocity Features: }}
Velocity features can be engineered to capture short-term behavioral bursts by aggregating transaction counts and amount sums over multiple rolling time windows per account. This helps detect rapid, high-frequency activity characteristic of fraud. These features should be withheld from the GNN node features to prevent temporal leakage through the message-passing neighborhood, but can be used during the downstream LightGBM stage (Refer Section~\ref{sec:lgbm}).
\begin{equation}
n_w = \sum_{t \in \mathcal{T}} \mathbf{1}[\tau - \tau_t \leq w],
\qquad
A_w = \sum_{t \in \mathcal{T}} \mathbf{1}[\tau - \tau_t \leq w] \cdot T_t
\end{equation}
where $w \in \{1\text{hr}, 6\text{hr}, 24\text{hr}\}$ denotes the rolling time window, $\mathcal{T}$ is the set of all prior transactions for the account, $\tau$ is the timestamp of the current transaction, $\tau_t$ is the timestamp of transaction $t$, and $T_t$ is the transaction amount at time $t$.

\item{\textbf{High-Dimensional Aggregate Features: }}
Financial datasets often anonymize some features for privacy-preserving reasons. If these are high-dimensional features, they can be condensed by selecting a subset of the most informative features and averaging them into a single risk-composite score. This can reduce dimensionality while preserving the aggregate signal these groups contribute to fraud detection.
\begin{equation}
f_{V\text{-risk}} = \frac{1}{|S_V|} \sum_{j \in S_V} V_j
\end{equation}
where $S_V$ is the selected subset of the feature group.

All the categorical fields in the dataset, such as credit card number or type, can be one-hot encoded.

\end{enumerate}

\item{\textbf{Account-Level Features: }}
Account identifier might already be present in the dataset, or can be constructed by concatenating card information per transaction and further grouping common accounts. Account-level features can be created using different aggregations of the individual transaction features connected to that account. Some meaningful transaction feature aggregations are described below.

\begin{equation}
\mathbf{a}_i =
\bigl[
\text{count}_i,\ 
\mu_i^{\text{amt}},\ 
\max_i^{\text{amt}},\ 
\Delta t_i^{\text{days}},\ 
\hat{r}_i
\bigr]
\end{equation}

where
\begin{align*}
\text{count}_i &: \text{number of transactions made from account } i \\
\mu_i^{\text{amt}} &: \text{mean transaction amount made from account } i \\
\max_i^{\text{amt}} &: \text{maximum transaction amount made from account } i \\
\Delta t_i^{\text{days}} &: \text{number of active days for account } i \\
\hat{r}_i &: \text{fraud rate for account (computed only for training set, to prevent data leakage)}
\end{align*}
\end{enumerate}
The features that are added to the transaction-level nodes correspond to the transaction feature vector, and the features that are added to the account-level nodes correspond to the account feature vector.

\subsubsection{Graph Construction}
The graph is formalized as $\mathcal{G} = (\mathcal{V}, \mathcal{E})$, where $\mathcal{V} = \mathcal{V}_{\text{acct}} \cup \mathcal{V}_{\text{txn}}$ is the union of account and transaction nodes, and $\mathcal{E}$ is a typed edge set. 

Four edge types are defined:
\begin{itemize}
\item \textbf{Account-Transaction edges}: A bidirectional edge connects each transaction node to its originating account node. These are the primary structural edges of the graph and carry LightGBM-derived fraud risk scores as edge weights (Note: the fraud risk score assignment is discussed in Section~\ref{sec:lgbm}).
\item \textbf{Shared Address edges}: Accounts sharing the same address field are connected to one another.
\item \textbf{Shared Email edges}: Accounts sharing the same email domain name are linked. Common email providers are excluded, as their connections are too broadly shared to carry identity signal and risk introducing noise into the graph.
\item \textbf{Shared Card edges}: Accounts sharing the same card type are connected, as it is a stronger identifier for a user than address or email for a credit card fraud detection problem.
\end{itemize}

Based on the graph construction strategy and entities involved, different edges are grouped by common attributes and a primary edge to hold the LightGBM-derived fraud risk scores can be decided. All edges are made bidirectional by including both $(u, v)$ and $(v, u)$ directions. The resulting graph is heterogeneous in both the node type and edge type, enabling the GNN to treat each relation separately during message passing.

Different node features with different dimensions are assigned for accounts and transactions which are independently standardized using \texttt{StandardScaler}, fit only on the training-split data. To unify them into a shared input space, both the feature vectors (the transaction-level and account-level features) are padded with zeroes to a width of $\max(d_{\text{txn}},\, d_{\text{acct}})$. A scalar degree feature corresponding to the normalized count of edges incident to each account node is appended to account node vectors, while a zero placeholder is appended to transaction node vectors at the same position.

\begin{figure}[ht]
\centering
\includegraphics[width=0.75\linewidth]{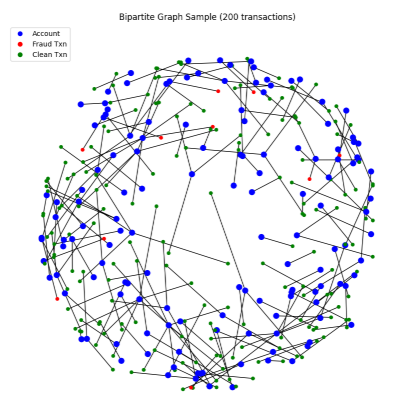}
\caption{A subgraph sampled over 200 transactions. Account nodes (blue) are connected to clean transactions (green) and fraudulent transactions (red) using account-transaction edges. Each edge depicts a transaction made using the account it is connected to.}
\label{fig:bipartite}
\end{figure}

The final bipartite structure of the graph is illustrated in Figure~\ref{fig:bipartite}.

\subsection{Risk Scoring with LightGBM}
\label{sec:lgbm_riskscoring}

Gradient boosting models are well established as strong probability estimators on tabular data and outperform deep learning models on structured feature sets and tabular data. RGH-GNN leverages this by using LightGBM as a first-stage risk scorer, producing transaction-level fraud probabilities that are passed to the GNN as edge weights.  This two-stage design allows the GNN to incorporate transaction-level fraud risk as a structural signal during message passing, rather than treating all edges as equally weighted.

\subsubsection{LightGBM Risk Scorer}

The LightGBM based risk scores are used to weight edges in the heterogeneous graph which allows the graph to learn relationships with context on both individual and relational transaction patterns. A LightGBM model is trained on raw tabular features that are independent to each transaction and distinct from the GNN node features. 

\subsubsection{Edge Weight Assignment and Thresholding}
\label{sec:lgbm}

Once trained, the model scores all transactions with a fraud probability $\hat{p}_i \in [0, 1]$ for each transaction $i$. These raw probabilities are assigned directly as edge weights on account-transaction edges in both directions, while cross-account edges (shared address, email, card) receive a uniform weight of 1.0:
\begin{equation}
w_i = \hat{p}_i, \quad e_i \in \mathcal{E}_{\text{acct-txn}}
\end{equation}

\begin{figure}[ht]
\centering
\includegraphics[width=0.75\linewidth]{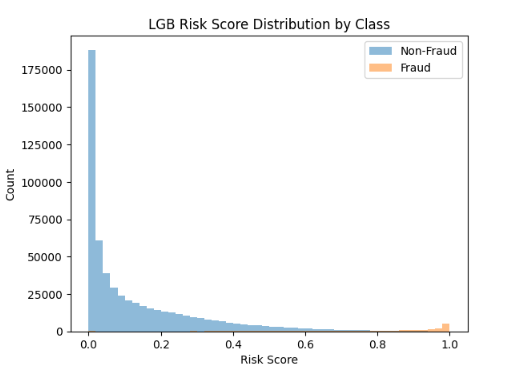}
\caption{Distribution of LightGBM risk scores by class across all 590,540 transactions. Non-fraud transactions (blue) concentrate near zero (mean score 0.131) while fraud transactions (orange) accumulate near 1.0 (mean score 0.793), a margin of 0.662 between classes, with the optimal classification threshold $\tau = 0.75$ separating the two populations.}
\label{fig:lgb_scores}
\end{figure}

These probability scores or risk scores are intentionally kept out of the transaction node features, since including them would create a direct mapping between the transaction label and node representation, bypassing the edge-guided relational learning the GNN is designed to perform. After training the LightGBM scorer, the distributions of the both classes were visualized as illustrated in Figure~\ref{fig:lgb_scores}.

\subsubsection{Model Architecture: RGH-GNN}
\begin{figure}[ht]
    \centering
    \includegraphics[width=1.15\textwidth]{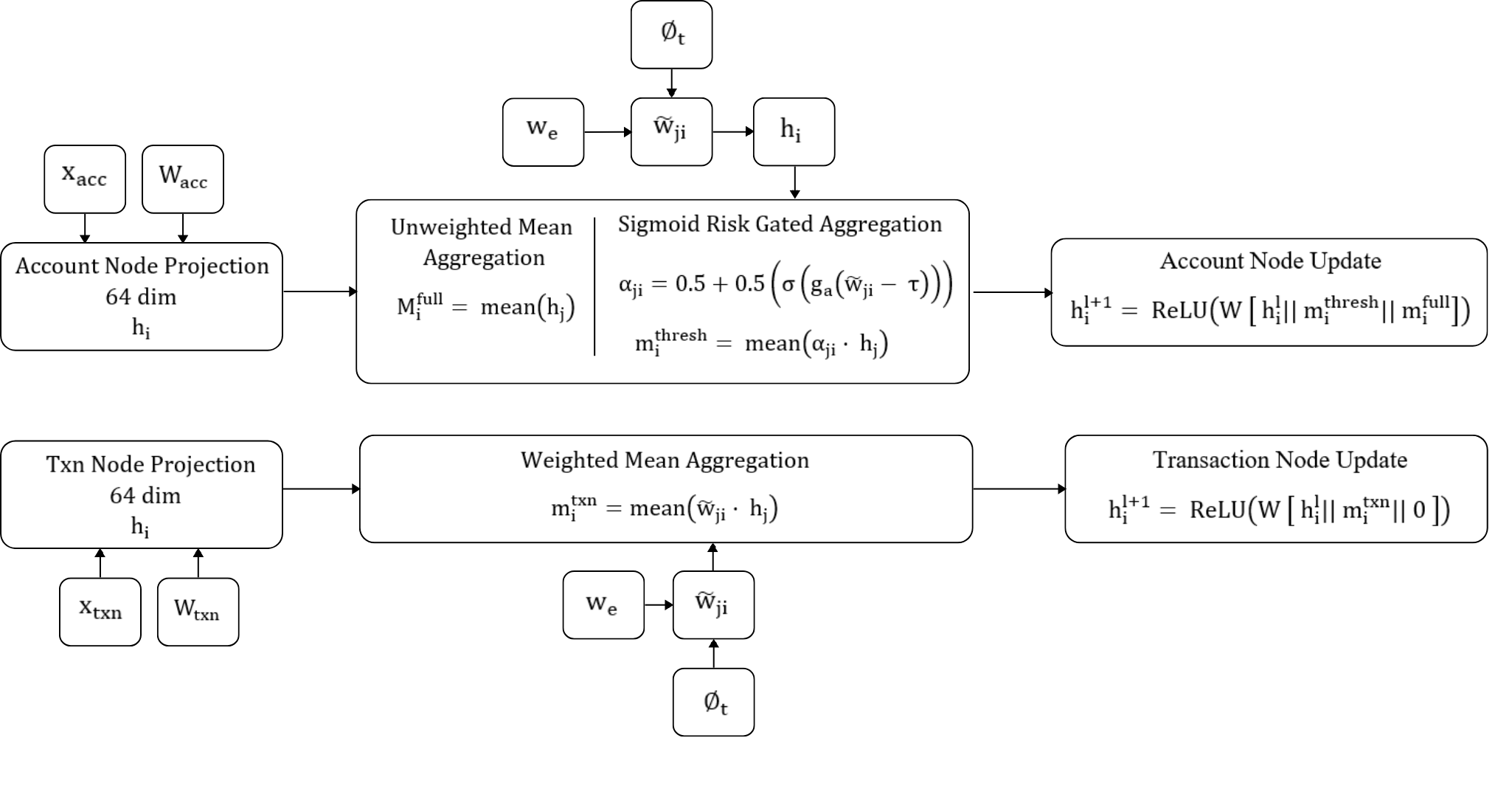}
    \caption{Forward-pass architecture of the proposed RiskCONV layer}
    \label{fig:forwardPass}
\end{figure}

As mentioned in the previous section, the risk scores have been generated and included as weights to the account-transaction edges of the graph. This is used as the input to the RGH-GNN algorithm proposed in the following section. RGH-GNN is a three-layer heterogeneous graph neural network built on a custom message passing layer, RiskConv, which includes edge-type awareness and risk-weighted aggregation. Figure~\ref{fig:forwardPass} illustrates the forward pass of the proposed RiskConv layer, showing the distinct aggregation mechanisms used for account and transaction nodes.

\subsection{GNN Training and Fraud Detection}

\subsubsection{Input Projection}

Account and transaction node features are semantically different, and hence projected separately into a shared 64-dimensional hidden space before message passing:

\begin{equation}
\mathbf{h}_i^{(0)} = \begin{cases}
\text{ReLU}(\mathbf{W}_{\text{acct}}\, \mathbf{x}_i) & \text{if } i \in \mathcal{V}_{\text{acct}}\\
\text{ReLU}(\mathbf{W}_{\text{txn}}\, \mathbf{x}_i) & \text{if } i \in \mathcal{V}_{\text{txn}}
\end{cases}
\end{equation}

This separation ensures that the model does not confuse account-level aggregations with transaction-level features.

\subsubsection{Edge Weights}

Each edge type is assigned a learnable scalar logit $\phi_t$, initialized to zero so that $\sigma(\phi_t) = 0.5$ at the start of training. The effective edge weight for edge $e$:

\begin{equation}
    \tilde{w}_e = w_e \cdot \sigma(\phi_t)
\end{equation}

where $w_e$ is the LightGBM-derived risk score for account-transaction edges or 1.0 for other edges. This allows the model to learn how much each node connecting the edge should contribute to the aggregation.

\subsubsection{Account Node Aggregation}

Account node aggregation is split across two pathways. For account-transaction edges, two aggregations are computed.

First, a full unweighted mean of all neighbouring transaction messages:

\begin{equation}
    \mathbf{m}_i^{\text{full}} = \operatorname{mean}_{j \in \mathcal{N}_0(i)} \mathbf{h}_j
\end{equation}

Second, a risk-gated weighted mean where each message is scaled by a shifted sigmoid gate:

\begin{equation}
    \alpha_{ji} = 0.5 + 0.5 \cdot \sigma\bigl(g_a \cdot (\tilde{w}_{ji} - \tau)\bigr)
\end{equation}

\begin{equation}
    \mathbf{m}_i^{\text{thresh}} = \operatorname{mean}_{j \in \mathcal{N}_0(i)} \alpha_{ji} \cdot \mathbf{h}_j
\end{equation}

where $\tau$ and $g_a$ are tuned hyperparameters that are assigned values 0.4 and 5.0 respectively. These values are selected by sweeping across an interval and selecting the configuration that provided the best validation performance (refer Table~\ref{tab:hyperparam-sweep}, Section~\ref{sec:setup}).
The shifted sigmoid floors the gate at 0.5 rather than near zero, ensuring that even low-risk transactions contribute some signal rather than being effectively masked out. Both the unweighted mean and risk-gated mean are computed in parallel, which helps preserve broader neighborhood context while also prioritizing discriminative aggregation.
For other types of edges, messages are aggregated via a standard unweighted mean.

\subsubsection{Transaction Node Aggregation}

Transaction nodes are given their parent account node aggregations scaled by their effective edge weight:

\begin{equation}
    \mathbf{m}_i^{\text{txn}} = \tilde{w}_{ji} \cdot \mathbf{h}_j
\end{equation}

\subsubsection{Node Update}

The final account node update concatenates the current node representation, the gated aggregation, and the full aggregation, then finally passes through a linear transformation:

\begin{equation}
    \mathbf{h}_i^{(l+1)} = \text{ReLU}\bigl(\mathbf{W}\,[\mathbf{h}_i^{(l)} \| \mathbf{m}_i^{\text{thresh}} \| \mathbf{m}_i^{\text{full}}]\bigr)
\end{equation}

The node update allows the model to see both the risk-filtered and unfiltered views of the neighborhood simultaneously, preventing the gating mechanism from discarding useful context. Similarly, transaction nodes are also updated based on their aggregation.
\paragraph{Classifier Head}

After three rounds of RiskConv, the transaction node representations are passed through a two-layer MLP classifier producing a single logit per transaction node. During inference, these logits are converted to probabilities via sigmoid, and a binary fraud prediction is made by applying the selected threshold (threshold selection procedure is outlined in Section~\ref{sec:training}).

\subsubsection{Loss Function and Class Imbalance}

Transaction fraud datasets exhibit severe class imbalance. IEEE-CIS has a fraud rate of approximately 3.5\%, which would heavily bias the model toward the majority class. RGH-GNN addresses this using BCEWithLogitsLoss with a pos\_weight multiplier that upweights the gradient contribution of fraud examples during training:

\begin{equation}
    \mathcal{L}_{\text{BCE}} = -\frac{1}{N}\sum_{i=1}^{N} \left[ w^+ \cdot y_i \log\sigma(\hat{z}_i) + (1 - y_i)\log\bigl(1 - \sigma(\hat{z}_i)\bigr) \right]
\end{equation}

where $w^+$ is the pos\_weight applied to the fraud class. An initial $w^+$ is set as the ratio of negative to positive training samples during initial training. A pos\_weight tuning sweep is then conducted over a range of multiplier values, each evaluated on validation F1 score, and the best $w^+$ is selected for final retraining (refer Table~\ref{tab:posweight-sweep} Section~\ref{sec:setup}).

\section{Experiments and Results}
\label{sec:experiments}

\subsection{Dataset}
The dataset chosen for this experiment is the IEEE-CIS Fraud Detection Dataset \cite{ieee_cis}. It includes both transactions and identities. The transaction dataset has 393 features with 17 named features and 376 anonymous features. The identity dataset has 41 features. The identity dataset has 144,234 instances and the transaction dataset has 590,540 instances. The transaction dataset contains features related to transaction-level attributes and the identity dataset contains features related to the identities that perform these transactions. 

\subsection{Experimental Setup}
\label{sec:setup}
The dataset is initally split into 70\% training and 30\% test set with stratification, preserving the original fraud rate of approximately 3.5\% across both partitions. Labels are assigned only to transaction nodes, with account nodes serving as unlabeled structural context for message passing. 

During the LightGBM Risk Scoring experiment, the 70\% training partition is further subdivided via a stratified 90/10 split, yielding approximately 63\% of the full dataset for model training and 7\% for validation. This split is deliberately carved out to prevent data leakage. Without it, risk scores assigned to the test set could reflect label information seen during training, which compromises the integrity of edge weights on unseen transactions. The validation set is held out during LightGBM training and used exclusively for early stopping and threshold selection.

RGH-GNN is trained and validated on the partition using the training procedure mentioned in Section~\ref{sec:training}. It has a novel risk aggregation strategy and is trained using a weighted loss function. The final evaluation and metrics are reported on the test set. 

Table~\ref{tab:hyperparam-sweep} below presents the grid search conducted to select the coefficients $\tau$ and $g_a$ for the gated-risk aggregation strategy and Table~\ref{tab:posweight-sweep} below presents the \texttt{pos\_weight} value picked for the weighted BCE loss function based on maximized validation F1 score. 

\begin{table}[ht]
\centering
\caption{Grid search results for $\tau$ and $g_a$, reporting validation F1 score. The best configuration ($\tau=0.4$, $g_a=5.0$) is shown in bold.}
\label{tab:hyperparam-sweep}
\begin{tabular}{ccc|ccc|ccc}
\toprule
$\tau$ & $g_a$ & F1 & $\tau$ & $g_a$ & F1 & $\tau$ & $g_a$ & F1 \\
\midrule
0.1 & 1.0  & 0.6318 & 0.2 & 20.0 & 0.6355 & 0.4 & 10.0 & 0.6315 \\
0.1 & 5.0  & 0.6360 & 0.3 & 1.0  & 0.6331 & 0.4 & 15.0 & 0.6376 \\
0.1 & 10.0 & 0.6352 & 0.3 & 5.0  & 0.6340 & 0.4 & 20.0 & 0.6336 \\
0.1 & 15.0 & 0.6359 & 0.3 & 10.0 & 0.6359 & 0.5 & 1.0 & 0.6313 \\
0.1 & 20.0 & 0.6303 & 0.3 & 15.0 & 0.6373 & 0.5 & 5.0 & 0.6340 \\
0.2 & 1.0  & 0.6378 & 0.3 & 20.0 & 0.6358 & 0.5 & 10.0 & 0.6315 \\
0.2 & 5.0  & 0.6334 & 0.4 & 1.0  & 0.6326 & 0.5 & 15.0 & 0.6362 \\
0.2 & 10.0 & 0.6274 & 0.4 & 5.0  & \textbf{0.6389} & 0.5 & 20.0 & 0.6350 \\
0.2 & 15.0 & 0.6350 &     &      &        &     &      &        \\
\bottomrule
\end{tabular}
\end{table}

\begin{table}[ht]
\centering
\caption{Validation F1 score across the pos\_weight ($w^+$) sweep. The best configuration ($w^+=1.5$) is shown in bold.}
\label{tab:posweight-sweep}
\resizebox{0.5\textwidth}{!}{%
\begin{tabular}{c c c c c c}
\toprule
$w^+$ & 1 & 1.5 & 2 & 3 & 5 \\
\midrule
F1 & 0.6286 & \textbf{0.6311} & 0.6203 & 0.6250 & 0.6160 \\
\bottomrule
\end{tabular}%
}
\end{table}
\subsection{Metrics}

The metrics used for evaluation are AUC-ROC, AUC-PR, F1, Precision, and Recall. AUC-ROC is the most widely reported metric in IEEE-CIS fraud detection literature. It is included here to show a comparison between baselines and their performance. AUC-PR is an important metric for settings with class imbalance as it directly captures model performance on the minority fraud class. Precision and Recall are included to provide a comparison on the trade-off between reducing false alarms and missing true positives at the fixed threshold. F1 score serves as the primary optimization target throughout training.

\subsection{Training and Validation Setup}
\label{sec:training}

RGH-GNN is trained on the heterogeneous transaction graph constructed in Section~\ref{sec:graph_construction}, using the training partition with a stratified 90/10 internal split for validation. Training proceeds in two stages. In the first stage, the model is trained on the training subset with evaluation on the validation subset every 20 epochs. At each evaluation step, the optimal classification threshold is selected by sweeping the precision-recall curve on the validation set and choosing the threshold that maximises F1 score. Early stopping is applied to save the best model state by validation F1 score. In the second stage, the best pos\_weight multiplier is identified via a separate sweep, and the model is retrained from scratch on the full training partition which includes training and validation combined. The best model state is saved and used for final test evaluation.

Each training stage for RGH-GNN and the subsequent models (ablation and baseline comparison models) were trained with the same strategy to maximize validation F1 score and prevent underfitting. Each model was trained for 2000 epochs with a patience counter of 5.

\subsection{Ablation Study}

RGH-GNN combines three architectural components that are its core contribution: heterogeneous node representation, LightGBM-derived risk scores as edge weights, and risk-gated aggregation. To quantify the individual contribution of each component, an ablation study was conducted where each component was individually removed while retaining the other two, and performance was measured. Following this, multiple components were removed simultaneously using GraphSAGE \cite{graphsage} as the base architecture.

The ablation configurations are as follows.

\begin{enumerate}
    \item \textbf{No-heterogeneity ablation:} Separate account and transaction input projections were replaced with a single shared projection applied to all nodes. The type-specific account and transaction aggregation rules were replaced with a uniform gated aggregation for both the node types. The risk scores and the edge-type relation weighting were preserved in the graph structure.
    \item \textbf{No-risk-scores ablation:} All edge weights were set to a uniform value of 1.0, removing all LightGBM-derived risk signal from the graph while keeping the full RiskConv architecture (including the gating mechanism) intact.
    \item \textbf{No-gated-aggregation ablation:} The shifted sigmoid gate aggregation mechanism during message passing was removed and replaced with a plain risk-weighted mean aggregation over account node neighbours. The node update was reduced to a single risk-weighted mean instead of the dual gated and full unweighted mean for account nodes. The risk scores on edges and node-type heterogeneity were preserved.
\end{enumerate}

Table~\ref{tab:ablation} reports the performance of each ablation variant alongside the full RGH-GNN model and Table~\ref{tab:ablation_components} summaries which components are present in each configuration.

\begin{table}[ht]
\centering
\caption{Ablation study results on the IEEE-CIS fraud detection test set.}
\label{tab:ablation}
\resizebox{\textwidth}{!}{%
\begin{tabular}{lccccc}
\toprule
\textbf{Model} & \textbf{AUC-ROC} & \textbf{AUC-PR} & \textbf{F1} & \textbf{Precision} & \textbf{Recall} \\
\midrule
\textbf{RGH-GNN (Full)} &
$\mathbf{0.914 \pm 0.001}$ &
$\mathbf{0.631 \pm 0.002}$ &
$\mathbf{0.619 \pm 0.002}$ &
$\mathbf{0.734 \pm 0.014}$ &
$\mathbf{0.536 \pm 0.009}$ \\
No Gate Aggregation & $0.912 \pm 0.001$ & $0.624 \pm 0.003$ & $0.616 \pm 0.003$ & $0.726 \pm 0.016$ & $0.535 \pm 0.009$ \\
No Risk Scores & $0.885 \pm 0.003$ & $0.399 \pm 0.017$ & $0.414 \pm 0.010$ & $0.400 \pm 0.029$ & $0.432 \pm 0.023$ \\
No Heterogeneity & $0.885 \pm 0.002$ & $0.407 \pm 0.013$ & $0.420 \pm 0.006$ & $0.396 \pm 0.010$ & $0.448 \pm 0.023$ \\
Weighted GraphSAGE & $0.879 \pm 0.003$ & $0.401 \pm 0.019$ & $0.418 \pm 0.015$ & $0.451 \pm 0.025$ & $0.392 \pm 0.035$ \\
Unweighted GraphSAGE & $0.884 \pm 0.001$ & $0.387 \pm 0.011$ & $0.405 \pm 0.008$ & $0.375 \pm 0.036$ & $0.445 \pm 0.035$ \\
\bottomrule
\end{tabular}%
}
\end{table}

\begin{table}[ht]
\centering
\caption{Component presence across ablation configurations.}
\label{tab:ablation_components}
\resizebox{\textwidth}{!}{%
\begin{tabular}{lccccc}
\toprule
\textbf{Model} & \textbf{\shortstack{Risk Scores}} & \textbf{Heterogeneity} & \textbf{\shortstack{Gated Aggregation}}\\
\midrule
\textbf{RGH-GNN (Full)} & \cmark & \cmark & \cmark \\
No Gated Aggregation     & \cmark & \cmark & \xmark \\
No Risk Scores           & \xmark & \cmark & \cmark \\
No Heterogeneity         & \cmark & \xmark & \cmark \\
Weighted GraphSAGE       & \cmark & \xmark & \xmark\\
Unweighted GraphSAGE     & \xmark & \xmark & \xmark \\
\bottomrule
\end{tabular}%
}
\end{table}

The ablation results demonstrate that all three components contribute positively to model performance, with the full RGH-GNN architecture achieving the best performance across all ablation configurations. Removing risk scores from the graph causes a drop in AUC-ROC from 0.914 to 0.885, confirming that the LightGBM-derived fraud probability signal is a critical input to the model. Removing heterogeneity produces a similarly substantial drop in AUC-ROC from 0.914 to 0.885, demonstrating that type-specific node representations (such as account and transaction nodes) and aggregation rules are important for extracting meaningful context in financial settings. A unified homogeneous representation, where different entities share the same feature set, is insufficient to capture the distinct characteristics of each entity type. Removing the gated aggregation also produces a marginal dip in the performance, as observed in~\ref{tab:ablation}. This indicates that the risk scores alone already provide a strong signal during aggregation, but the gate ensures that high-risk transactions contribute more than low-risk ones. This aggregation strategy is essential to ensuring that the risk scores are utilized properly during message passing.

The comparison between Weighted and Unweighted GraphSAGE, in~\ref{tab:ablation} confirms that the risk scores alone provide meaningful signal to improve the performance. However, it requires both heterogeneous node treatment and the gated aggregation mechanism to properly exploit them.

\subsection{Baselines}

We compare RGH-GNN against a range of graph neural network baselines: GraphSAGE \cite{graphsage}, GAT \cite{gat}, GIN \cite{gin}, CARE-GNN \cite{care_gnn}, PC-GNN \cite{pcgnn}, HAN \cite{han}, and HGT \cite{hgt}. CARE-GNN and PC-GNN were originally designed for homogeneous multi-relation graphs, in which all relations connect nodes of the same type, and rely on reinforcement-learning components for neighbor selection. This cannot be directly applied to our bipartite heterogeneous transaction graph. To enable a meaningful comparison, we instead extract the core components of each architecture and implement them on top of the GraphSAGE architecture: the similarity-aware neighbor filtering mechanism from CARE-GNN \cite{care_gnn}, and the label-balanced neighbor sampler from PC-GNN \cite{pcgnn}, are implemented and compared against RGH-GNN.

For most of the baselines, we report two variants:

\begin{enumerate}
    \item \textbf{Unweighted}: This variant runs each model on the heterogeneous transaction graph using the same node features and experimental procedure described in Section~\ref{sec:training}. However, the graph sent as input to these models does not contain the engineered risk scores. This allows us to identify the contribution of risk-score integration in the RGH-GNN pipeline and how it improves the performance.
    \item \textbf{Weighted}: This variant includes the risk-score edge weights as input to each baseline, to compare whether each architecture can exploit this additional signal as effectively as RGH-GNN.
\end{enumerate}

\subsection{Results}
\begin{table}[ht]
\centering
\caption{Performance comparison of baseline models against RGH-GNN on the IEEE-CIS fraud detection test set. The best performance is highlighted in bold and the second-best is underlined.}
\label{tab:results}
\resizebox{\textwidth}{!}{%
\begin{tabular}{lccccc}
\toprule
\textbf{Model} & \textbf{AUC-ROC} & \textbf{AUC-PR} & \textbf{F1} & \textbf{Precision} & \textbf{Recall} \\
\midrule
Unweighted GraphSAGE \cite{graphsage} & $0.884 \pm 0.001$ & $0.387 \pm 0.011$ & $0.405 \pm 0.008$ & $0.375 \pm 0.036$ & \underline{$0.445 \pm 0.035$} \\
Weighted GraphSAGE \cite{graphsage}& $0.879 \pm 0.003$ & \underline{$0.401 \pm 0.019$} & \underline{$0.418 \pm 0.015$} & \underline{$0.451 \pm 0.025$} & $0.392 \pm 0.035$ \\
Unweighted GIN \cite{gin}& $0.734 \pm 0.002$ & $0.146 \pm 0.009$ & $0.242 \pm 0.011$ & $0.182 \pm 0.017$ & $0.368 \pm 0.026$ \\
Weighted GIN \cite{gin}& $0.642 \pm 0.031$ & $0.073 \pm 0.004$ & $0.186 \pm 0.008$ & $0.130 \pm 0.010$ & $0.332 \pm 0.021$ \\
Unweighted GAT \cite{gat}& $0.862 \pm 0.003$ & $0.319 \pm 0.009$ & $0.363 \pm 0.004$ & $0.363 \pm 0.036$ & $0.367 \pm 0.030$ \\
Weighted GAT \cite{gat}& $0.865 \pm 0.004$ & $0.331 \pm 0.020$ & $0.372 \pm 0.016$ & $0.371 \pm 0.024$ & $0.374 \pm 0.015$ \\
Unweighted HAN \cite{han}& $0.811 \pm 0.007$ & $0.170 \pm 0.008$ & $0.272 \pm 0.005$ & $0.217 \pm 0.005$ & $0.363 \pm 0.009$ \\
Unweighted HGT \cite{hgt}& $0.624 \pm 0.001$ & $0.177 \pm 0.000$ & $0.269 \pm 0.001$ & $0.260 \pm 0.000$ & $0.278 \pm 0.000$ \\
Unweighted CARE-GNN* \cite{care_gnn}& \underline{$0.885 \pm 0.003$} & \underline{$0.401 \pm 0.017$} & $0.416 \pm 0.012$ & $0.421 \pm 0.025$ & $0.413 \pm 0.016$ \\
Weighted CARE-GNN* \cite{care_gnn}& \underline{$0.885 \pm 0.004$} & $0.399 \pm 0.016$ & $0.415 \pm 0.009$ & $0.410 \pm 0.031$ & $0.424 \pm 0.016$ \\
Unweighted PC-GNN* \cite{pcgnn}& $0.623 \pm 0.002$ & $0.074 \pm 0.001$ & $0.154 \pm 0.001$ & $0.109 \pm 0.002$ & $0.264 \pm 0.012$ \\
Weighted PC-GNN* \cite{pcgnn}& $0.653 \pm 0.001$ & $0.079 \pm 0.001$ & $0.163 \pm 0.000$ & $0.111 \pm 0.001$ & $0.309 \pm 0.010$ \\
\midrule
\textbf{RGH-GNN} &
$\mathbf{0.914 \pm 0.001}$ &
$\mathbf{0.631 \pm 0.002}$ &
$\mathbf{0.619 \pm 0.002}$ &
$\mathbf{0.734 \pm 0.014}$ &
$\mathbf{0.536 \pm 0.009}$ \\
\bottomrule
\end{tabular}%
}

\vspace{2pt}
{\footnotesize 
\raggedright
*CARE-GNN and PC-GNN denote only the core extracted components (similarity-aware neighbor filtering and label-balanced neighbor sampler, respectively) implemented on top of GraphSAGE, not the full original architectures.
\par}
\end{table}

Table~\ref{tab:results} presents the performance of all models on the held-out test set. RGH-GNN achieves the strongest overall performance across all evaluated models. This performance gain suggests that the explicit assignment of risk scores, and their incorporation during message-passing aggregation, allows the model to prioritize the most suspicious interactions rather than treating all neighboring transactions uniformly. The other baseline architectures do not consider this explicitely while aggregating neighborhood information. This indicates that risk-guided aggregation is a meaningful contributor to overall performance gains.

\subsection{Performance Analysis}

Among the twelve GNN baselines evaluated, RGH-GNN achieves the strongest overall performance. Both variants of CARE-GNN \cite{care_gnn} and GraphSAGE \cite{graphsage} achieve the next-best performance. CARE-GNN achieves an AUC-ROC of 0.885. Weighted GraphSAGE achieves an F1 score of 0.418 and a Precision of 0.451. Unweighted GraphSAGE achieves a Recall of 0.445.

Across these architectures, the gap between weighted and unweighted variants is generally small and inconsistent in direction. In the GIN \cite{gin} architecture, the weighted variant performs worse than the unweighted counterpart. Only PC-GNN \cite{pcgnn} shows a consistent improvement when risk scores are included across all its metrics. This suggests that most baseline architectures are not equipped to leverage and learn from the additional risk-score information even when they are provided as weights to the input graph. RGH-GNN's higher performance suggests that its gated aggregation and heterogeneity-aware design converts this risk-score information into a meaningful signal that it can learn to flag fraud with, far more effectively than any baseline tested.

A substantial margin is observed in AUC-PR, where RGH-GNN achieves 0.631 against the next-best performance of 0.401. AUC-PR is an important metric for settings with severe class imbalance, as it concentrates entirely on the fraud (minority) class and penalises models that gain recall at the expense of excessive false positives \cite{davis2006relationship}. The magnitude of this gap indicates that RGH-GNN produces substantially better-calibrated fraud probability estimates, rather than merely achieving higher ranking performance.

F1 score shows a similarly large improvement, rising from 0.418 (Weighted GraphSAGE) to 0.619 (RGH-GNN). All hyperparameters, including decision threshold, early stopping, and posweight, were selected during training by maximising validation F1 score. Therefore, this metric directly reflects the quality of each model's fixed operating point, rather than its ranking ability alone. The magnitude of this improvement suggests that the relational context present in the heterogeneous graph, through shared accounts, addresses, email domains, and card identifiers, provides a discriminative signal that RGH-GNN's risk-weighted aggregation exploits more effectively than the baseline architectures. AUC-ROC improves from 0.885 (CARE-GNN) to 0.914 (RGH-GNN), consistent with this metric being less sensitive to minority-class performance under heavy class imbalance. 
\section{Discussion}
\label{sec:discussion}
\begin{figure}[ht]
    \centering
    \includegraphics[width=0.75\linewidth]{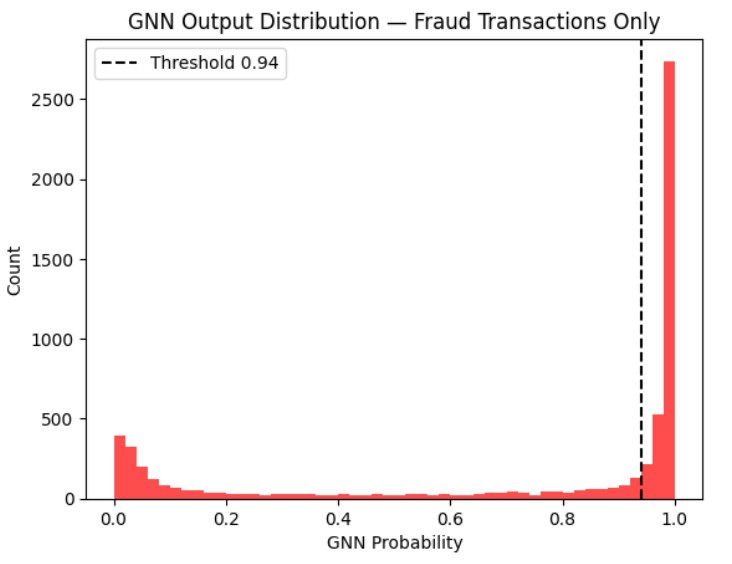}
    \caption{GNN output probability distribution for fraud transactions on the held-out test set. The dashed line marks the classification threshold at 0.94. The majority of fraud transactions accumulate near 1.0 and are correctly classified, while a cluster near 0 represents hard negatives (transactions the model confidently assigned low fraud probability despite being genuinely fraudulent.)}
    \label{fig:gnn_dist}
\end{figure}

The performance metrics demonstrate that RGH-GNN effectively generates and leverages risk scores for fraud detection. The proposed methodology allows financial systems to be represented as a heterogeneous graph while preserving entity-specific features for different entity types involved in transactions. The risk signals generated by the LightGBM classifier provide additional information during graph message passing, enabling more informed neighbour selection while incorporating both tabular and relational views into the fraud detection process. The gated aggregation mechanism further ensures that the risk scores are transformed into meaningful contextual information that can be leveraged for fraud detection, which some conventional algorithms struggle to do. RGH-GNN's output probability distribution for the fraud class, shown in Figure~\ref{fig:gnn_dist}, further indicates that its performance is primarily attributed to its ability to identify a majority of the existing fraudulent samples, rather than exploiting the high class imbalance by predicting multiple false-positives.

A key limitation of this methodology is revealed by the GNN output probability distribution for fraud transactions, shown in Figure~\ref{fig:gnn_dist}. While the majority of fraud transactions are correctly pushed above the 0.94 threshold, a small cluster of fraud transactions accumulates near zero. These are transactions the model assigned very high clean probability to, despite being genuinely fraudulent. The cluster appears only at the extreme low end of the probability range, with very little volume in the intermediate range between 0.1 and 0.9. This distribution suggests these are not cases of model uncertainty but rather transactions that the model was confidently wrong about. This implies that the cluster of transactions are structurally distinct from the fraud cases the model successfully identifies, potentially representing highly engineered fraudulent transactions that mimic legitimate behaviour across all the recorded observable features and relationships in this pipeline. Isolating and studying these transactions, understanding what patterns characterise them, and identifying what new features or views might flag them is a meaningful direction for future work (similar to how relational context was found here to add signal beyond individual features alone).

A second limitation concerns dataset availability. The IEEE-CIS dataset is a valuable public benchmark, but publicly available financial transaction datasets with sufficient scale and label quality for graph-based fraud detection research remains scarce. A majority of real-world transaction data is proprietary. This restricts the ability to test and derive methodologies across diverse fraud patterns. It also limits the ability to directly compare methodology output with prior work. Very few graph-based approaches on IEEE-CIS dataset report the full range of metrics used in this work, making external benchmarking difficult. Expanding evaluation to additional public datasets and more standardised performance reporting based on meaningful metrics for fraud detection literature is also an area of scope to explore further.

\section{Conclusion}
\label{sec:conclusion}

This work presents RGH-GNN, a supervised graph-based framework for financial fraud detection that uses risk signals to guide message passing. A LightGBM model, chosen for its strong performance on tabular classification, generates transaction-level fraud risk scores that are used to weight edges in a heterogeneous account-transaction graph. A learnable gating mechanism, tuned to maximise validation performance, ensures that higher-risk transactions contribute more strongly to the final classification. This allows the model to address both the camouflage problem, where fraudsters disguise transactions to resemble legitimate behaviour and the severe class imbalance which is a known problem in financial fraud datasets. Distinct node representations for accounts and transactions allow the heterogeneous structure of the graph to capture meaningful contextual information that a homogeneous representation would fall short at.

RGH-GNN is evaluated against standard graph neural network baselines, including uniform-aggregation and heterogeneous-aware architectures. These are trained on the same heterogeneous graph and node features similar to the proposed pipeline. RGH-GNN outperforms all baselines across every metric, demonstrating that explicitly incorporating risk signals into message passing improves the model's ability to flag fraud by more effectively addressing camouflage and class imbalance.

A key limitation is a cluster of outlier fraud transactions that bypass detection even under this framework, suggesting structurally distinct fraud patterns not captured by the current feature set or graph structure. Future work includes identifying and modeling features distinct to this outlier distribution and to build more robust detection pipelines. Extending risk-aware modeling to settings where labelled data is scarce or unavailable, since proprietary financial data is rarely public, is another direction to explore. Exploring risk-aware graph representations in other financial domains such as supply chain finance and anti-money laundering networks, where the structurally relevant features and relationships differ from credit card fraud is also another meaningful direction to extend this work.

\section*{Acknowledgement}
This work was carried out at the Center for Cybersecurity, Shiv Nadar University Chennai, supported by the Shiv Nadar Foundation.

\bibliography{reference}

\end{document}